\documentclass[11pt,a4paper]{article}
\usepackage[hyperref]{emnlp-ijcnlp-2019}
\usepackage{times}
\usepackage{latexsym}
\usepackage{amsmath}

\usepackage{graphicx}
\usepackage{multirow}
\usepackage{booktabs}
\usepackage{todonotes}

\usepackage{url}

\usepackage[font=small]{caption}
\aclfinalcopy 

\title{Simple yet Effective Bridge Reasoning \\ for Open-Domain Multi-Hop Question Answering}

\author{
 Wenhan Xiong$^\dagger$,
 Mo Yu$^\ddagger$, 
 Xiaoxiao Guo$^\ddagger$, 
 Hong Wang$^\dagger$,
 Shiyu Chang$^\ddagger$, \\
 \textbf{
 Murray Campbell$^\ddagger$,
 William Yang Wang$^\dagger$}
\\ 
 $^\dagger$ University of California, Santa Barbara\\
 $^\ddagger$ IBM Research\\
 \{xwhan, william\}@cs.ucsb.edu,\\ \{yum,mcam\}@us.ibm.com, \{shiyu.chang, xiaoxiao.guo\}@ibm.com  
 }

\date{}

\begin{document}
\maketitle
\begin{abstract}
A key challenge of multi-hop question answering (QA) in the open-domain setting is to accurately retrieve the supporting passages from a large corpus. Existing work on open-domain QA typically relies on off-the-shelf information retrieval (IR) techniques to retrieve \textbf{answer passages}, i.e., the passages containing the groundtruth answers.
However, IR-based approaches are insufficient for multi-hop questions, as the topic of the second or further hops is not explicitly covered by the question. To resolve this issue, we introduce a new sub-problem of open-domain multi-hop QA, which aims to recognize the bridge (\emph{i.e.}, the anchor that links to the answer passage) from the context of a set of start passages with a reading comprehension model. This model, the \textbf{bridge reasoner}, is trained with a weakly supervised signal and produces the candidate answer passages for the \textbf{passage reader} to extract the answer. On the full-wiki HotpotQA benchmark, we significantly improve the baseline method by 14 point F1. Without using any memory-inefficient contextual embeddings, our result is also competitive with the state-of-the-art that applies BERT in multiple modules.

\end{abstract}

\section{Introduction}

As machines have achieved super-human performance~\cite{devlin2018bert} for single-passage question answering on the standard SQuAD dataset~\cite{rajpurkar2016squad}, building QA systems with human-like reasoning ability has attracted broad attention recently. In this challenge, the QA system is required to reason with distributed piece of information from multiple passages to derive the answer. Several multi-hop QA benchmarks include \textsc{WikiHop}~\cite{welbl2018constructing}, ComplexWebQuestions~\cite{talmor2018web} and HotpotQA~\cite{yang2018hotpotqa} have been released recently to advance this line of research. In this paper, we focus on the practical open-domain HotpotQA benchmark where the questions are asked upon natural language passages instead of knowledge bases and the supporting passages are not known beforehand.

The typical pipeline of open-domain QA systems~\cite{chen-etal-2017-reading,DBLP:conf/aaai/WangYGWKZCTZJ18,htut2018training} is to first use an IR system to retrieve a compact set of paragraphs and then run a machine reading model over the concatenated or reranked paragraphs. While IR works reasonably well for simple questions\footnote{As shown in Table 3 of \citet{chen-etal-2017-reading}, a simple IR method can achieve $77.8\%$ recall on the SQuAD dataset.}, it often fails to retrieve the correct answer paragraph for multi-hop questions. This is due to the fact that the question often cannot fully cover the information for the second or further hops. Consider the question \textit{``What government position was held by the woman who portrayed Corliss Archer in the film Kiss and Tell?"} from the HotpotQA~\cite{yang2018hotpotqa} dataset. 
Since the name of the person (\textit{Shirley Temple}) is not directly mentioned in the question and the answer is about another aspect of the person other than film acting, traditional IR heuristics based on $n$-gram matching might fail to retrieve the answer passage. In fact, the correct answer passage of \textit{Shirley Temple} never appears in the top passages ranked by the default IR method of HotpotQA. 



Instead of predicting the answer passage with text matching between passages and questions, we claim that the answer passage can be better inferred based on the context-level information. 
Noticing that the IR retrieved passages can usually successfully cover the first hop evidence of the questions (i.e. \textbf{start passages}), we propose to use a reading comprehension model to infer the entities linking to the answer passage from the start passages. Our experiments show that this simple approach can tremendously increase the answer coverage of the top-ranked passages and thus increase the final QA performance by 14 point F1. Despite that our bridge reasoner and passage reader only learn above GloVe embeddings~\cite{pennington2014glove}, we achieve competitive performance with methods that use BERT~\cite{devlin2018bert} in multiple modules.

\begin{figure}[t]
\centering
\includegraphics[width=1\linewidth]{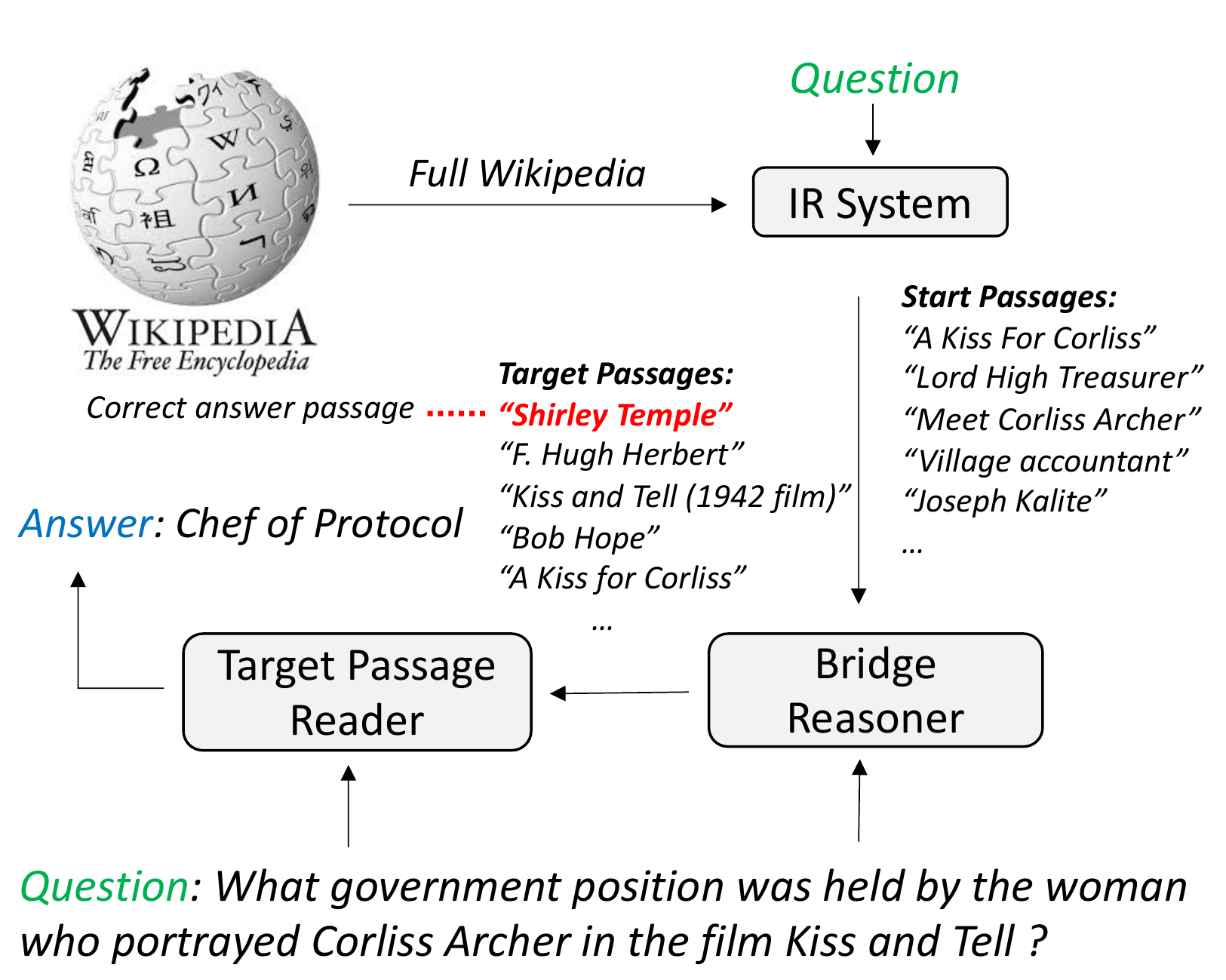}
\caption{The overview of our QA system. The \textbf{bridge reasoner} reads the start passages retrieved by an IR system and predicts a set of candidate bridges (anchor links) that lead to the answer passages, which is further processed by the \textbf{passage reader} to return the answer.}
\label{system}
\vspace{-0.2in}
\end{figure}

\section{Problem Definition and Motivation}
\label{sec:definition}
An open-domain multi-hop QA system aims to answer complex questions by retrieving evidence from a large open-domain passage corpus, such as Wikipedia. Usually the evidence scatters in a distributed set of supporting passages $p_1, p_2, ..., p_n$ that forms an ordered chain. Each $p_i$ provides evidence that partially fulfills the information required to answer the question, as well as provides clues (usually concepts or entities) that lead to the next supporting passage $p_{i+1}$. The last passage $p_n$ of the chain contains the answer and is referred to as the \textbf{answer passage}. 
%
Although the supervision of the complete supporting chains could be beneficial for training and diagnosing the QA system, predicting these complete reasoning sequences at evaluation time is usually quite challenging.

This work builds on an important observation that the prediction of the entire chain is not necessary for the QA performance. As a matter of fact, we conduct a preliminary experiment that compares a QA model that has full access to the supporting passages, versus a model that only has access to the answer passage. 
This experiment was conducted on the distractor version of HotpotQA~\cite{yang2018hotpotqa}, which has groundtruth supporting passage annotations. We use the baseline QA model from \citet{yang2018hotpotqa}.
The result shows that the full access only gives marginal improvements\footnote{66.07 F1 and 49.43 EM with full support access versus 64.77 F1 and 50.96 EM with only answer passage access.}, even this model uses the supporting passage labels as additional supervision signals.
The above result confirms that the multi-hop QA performance largely depends on the accurate retrieval of the answer passages.

\paragraph{Definition of Bridge Reasoning}
The key idea of our approach is to reformulate the problem of answer passage retrieval as a reading comprehension task. The reading model predicts an entity that points to the answer passage. Such entities serve as the bridges connecting the supporting passages, therefore 
we refer them as bridge entities. When working with passages from Wikipedia, we consider the anchor links in each article as the candidate set of bridge entities. Thus each bridge candidate is a title of another passage and 
 we use \textbf{bridge entity} and \textbf{answer passage} interchangablely.

Note that our definition of bridge reasoning here can be easily extended beyond anchor links, as long as we have entity linking tools to connect the same entities in different passages and build links between them. The main goal of this paper is to demonstrate that the bridge reasoning task can be effectively formulated as a reading comprehension task, and we leave the investigation of the broader definition of bridge reasoning to future work.

\paragraph{Remark on Distant Supervision} It is also worthy to note that obtaining the supervision of the answer passages is much easier -- as long as there are question-answer pairs, we can use distant supervision to obtain answer passage annotations. Therefore the proposed bridge reasoning task is rather general and is easy to be extend to more datasets without support passage supervision.


\section{The Proposed Approach}
Our QA system is illustrated in Figure~\ref{system}. We first use the \textbf{bridge reasoner} to get the answer passages and then feed the top candidate answer passages into a standard \textbf{passage reader} to predict the final answer to the multi-hop question.

\subsection{The Base Span Prediction Model}
Both the bridge reasoner and the passage reader use a model that predicts a relevant span given a question. We use the same model architecture for both tasks and the architecture is base on the document QA model from~\cite{clark2017simple}, which is used by~\citet{yang2018hotpotqa} as the baseline for HotpotQA. The model uses a shared bidirectional GRU~\cite{cho2014learning} to encode the question and the passages. The encoded questions and passages are then passed to a bidirectional attention layer~\cite{seo2016bidirectional} to get the question-aware passage states. The state vectors are enhanced by a self-attention layer~\cite{wang2017gated} and are finally fed into linear layers to predict the start and end span scores at every word position. 

\vspace{-0.05in}
\subsection{Bridge Reasoner}

Our bridge reasoner integrates multiple types of evidence to predict the bridge entities that link to potential answer passages.

\vspace{0.05in}
{\bf \noindent Local Context Evidence}
The most critical evidence we use is the local context of the start passages. These passages usually cover the first hop of the question and provide clues about the bridges. Our bridge reasoner therefore employs the span prediction model to predict the spans of bridge entities from the context of the start passages. Unlike typical span prediction models that consider all possible spans, the bridge reasoner here only needs to rank all the entities that have anchor links. We take the final representation of each token from the span prediction model and use each anchor's start token representation $h^c_{a_s}$ to represent the anchor's local context evidence.


{\bf \noindent Passage Content Evidence}
Each bridge entity in our setting is associated with a Wikipedia article, so  the relevance of each bridge can be computed by matching the article content with the question. Here we use a bi-LSTM to encode the abstract passages and use max-pooling on the output states to get the passage content representation $h^p_a$.




Both the local context evidence $h^c_{a_s}$ and passage content evidence $h^p_a$ are integrated into our final bridge reasoner by a linear layer. The supervision for training the bridge reasoner is derived from the distractor version of HotpotQA: we take the title of the support passage that contains the groundtruth answer as the groundtruth bridge entity. When there are multiple passages that contain the answer, we randomly pick one of the passages. 



\subsection{Target Passage Reader}
Our passage reader has the same neural architecture as the bridge reasoner and the goal here is to extract the correct answer span. We run the target passage reader on the top 10 answer passage candidates predicted by the bridge reasoner. 

\paragraph{Training Passages from Cross-validation}
As we are using the same set of training questions for training the bridge reasoner and the target passage reader, there will be a discrepancy between the training and evaluation of QA: at evaluation time, the reader sees the passages predicted by the bridge reasoner, while at training time, the groundtruth answer passage is known. On the other hand, we also cannot use the predicted passages for training the reader, as the bridge reasoner itself is trained on the training set so the top predicted passages on training set are already overfitted. To make the training match the evaluation, we use the bridge reasoner model to perform two-fold cross-validation on training questions and use the cross-predicted passages for training the reader. 

\paragraph{Auxiliary Training Objective of Bridge Prediction} We introduce an auxiliary objective to encourage the reader to utilize the answer passage supervision during training. This is  done by adding a span loss for predicting the answer passage title\footnote{The passage titles are included as part of the context for QA.}. This simple auxiliary loss introduces implicit regularization for the reader and turns to be beneficial for the final QA performance.




\section{Experiments}
\paragraph{Setup} Our experiments mainly focus on the ``\textit{bridge}" questions of which the supporting passages can form a reasoning chain and the answers can be found in the last passage. For the ``\textit{comparison}" questions in the dataset, the topics for comparison are often explicitly mentioned in the questions, so IR methods are often sufficient and we keep the IR retrieved passages for comparison answer prediction. Because HotpotQA does not provide training passages for the open-domain setting, we use a hybrid tf-idf and bm25 approach to retrieve 10 start passages for each training question. For the dev and test questions, we directly run the trained bridge reasoner on the start passages retrieved by HotpotQA's default IR approach. To further expand the coverage of the start passages, we find a useful external entity linking tool\footnote{\url{https://tagme.d4science.org/tagme/}} and we append the abstracts of the Top2 returned Wikipedia articles for both bridge reasoning and answer prediction.

\paragraph{Answer Passage Prediction} The performance of the bridge reasoner on answer passage predictions is shown in Table~\ref{tab:bridge_prediction}. Overall, the bridge reasoner retrieves the answer passage with significantly higher accuracy  
than HotpotQA's IR method. We also see that the local context evidence is more effective than the passage content evidence for answer passage prediction. Since conventional IR methods also use passage content for ranking, the results here validate our assumption that the bridges can be better inferred by reading the context of the start passages.

\begin{table}[t]
\small
    \centering
    \begin{tabular}{l|c}
    \toprule
       Approach & Hits@10\\
      \midrule
       HotpotQA IR & 48.4\\
    \midrule
    \multicolumn{2}{c}{\underline{\emph{Our Methods}}} \\
       Bridge Reasoner & 76.6\\
       \quad w/o local context evidence & 75.4 \\
       \quad w/o passage content evidence & 65.7\\
       Bridge Reasoner + entity linking & \bf 80.6\\
    \bottomrule
    \end{tabular}
    \caption{Answer passage prediction performance, measured by Hits@10 on dev bridge questions.}
    \label{tab:bridge_prediction}
\end{table}


\begin{table}[t]
\small
    \centering
    \begin{tabular}{l|cc|cc}
    \toprule
        \multirow{2}{0.8cm}{\textbf{Model}} & \multicolumn{2}{c|}{\textbf{Dev}} &\multicolumn{2}{c}{\textbf{Test}}\\
        & EM & F1 & EM & F1 \\
    \midrule
        \multicolumn{5}{c}{\emph{Methods w/o BERT}} \\
        HotpotQA Baseline & 24.68 & 34.36 & 23.95 & 32.89 \\
        GRN & - & - & 27.34 & 36.48 \\
        \textbf{Ours} & 36.81 & 48.48 & 36.04 & 47.43\\ 
        \quad w/o EL & \underline{35.00} & \underline{46.16} & - & -\\
    \midrule
     \multicolumn{5}{c}{\emph{Methods with BERT}} \\
        GRN + BERT & - & - & 29.87 & 39.14 \\
        CogQA & 37.6 & 49.4 & 37.12 & 48.87 \\
        \quad w/o EL & \underline{34.6} & \underline{46.2} & - & - \\
        \quad w/o re-scoring & 33.6 & 45.0  & - & -\\
    \midrule
        \multicolumn{5}{c}{\emph{Methods with Unknown Usage of BERT}} \\
        DecompRC & - & - & 30.00 & 40.65 \\
        MUPPET & - & - & 30.61 & 40.26 \\
    \bottomrule
    \end{tabular}
    \caption{QA performance on HotpotQA. The \underline{underline} methods use the same resource, but our method does not use any pre-trained contextual embeddings like BERT.}
    \label{tab:QA}
\end{table}

\paragraph{Question Answering Results} Table~\ref{tab:QA} shows the final multi-hop QA performance. We compare several concurrent systems on the leaderboard, 
including the newly published CogQA~\cite{ding2019cognitive} and a few anonymous results that are released at the same period as CogQA, \emph{e.g.}, MUPPET, GRN, and DecompRC. Most of the top systems on the leaderboard benefit from the pre-trained contextual embedding BERT, while our method is trained from scratch.
We categorize all the systems according to their usages of BERT. Among all the results without BERT, our approach shows a huge advantage and is about 10\% higher in terms of both EM and F1 compared to the current known best system w/o BERT (GRN). Since our reader has the same architecture as the HotpotQA baseline, this shows the great potential of our bridge reasoner. When compared to models w/ BERT, \emph{i.e.}, the CogQA, our result is still competitive. Similarly to CogQA, we also investigate the passage initialization with question entity linking, and observed significant performance boost. Note that the CogQA paper does not provide details of the entity linker, so the results with our entity linker may not be the same to the one used by CogQA. Furthermore, when entity linking is not used, our method and CogQA start with the same initial passages. This gives an apple-to-apple comparison except that ours does not use BERT. According to the dev results, our method is on par with CogQA (35.0 v.s. 34.6 for EM and both 46.2 for F1). This proves that our bridge reasoning method is superior to the cognitive graph generator in CogQA.

\paragraph{Ablation Study}
Table~\ref{tab:ablation} gives ablation results on the dev set, where both entity linking and the auxiliary objective slightly improve the performance. As the focus of the paper is to improve the coverage of answer passages for ``bridge" questions, we also report the ``bridge" question portion where the improvement is more significant.

\begin{table}[t]
\small
    \centering
    \begin{tabular}{l|cc|cc}
    \toprule
    \multirow{2}{0.5cm}{\textbf{Model}} & \multicolumn{2}{c|}{\bf Bridge Only} &  \multicolumn{2}{c}{\bf Full Dev}\\
     & \bf EM & \bf F1 & \bf EM & \bf F1 \\
    \midrule
        Our approach & 34.19 & 47.16 & 36.81 & 48.48 \\ 
        \quad w/o EL & 32.91 & 45.42 & 35.00 & 46.16 \\
        \quad w/o Multi-Task & 32.91 & 46.13 & 35.80 & 47.14\\
        \quad w/o Bridge Reasoner & 22.52 & 32.78 & 27.05 & 36.67\\
    \bottomrule
    \end{tabular}
    \vspace{-1ex}
    \caption{QA performance ablation on the development set.}
    \label{tab:ablation}
    \vspace{-0.2in}
\end{table}

\section{Conclusion}
This paper introduces an important sub-problem of bridge reasoning for the task of multi-hop QA in the open-domain setting. We propose a bridge reasoner that utilizes multiple types of evidence to derive the passages that cover the answers. The reasoner significantly improves the coverage of answer passages than IR methods. With the predicted passages, we show that a standard reading comprehension model is able to achieve similar performance as the state-of-the-art method that requires BERT in multiple modules.

\bibliography{emnlp-ijcnlp-2019}
\bibliographystyle{acl_natbib}

\appendix

\end{document}